# IslamicPCQA: A Dataset for Persian Multi-hop Complex Question Answering in Islamic Text Resources

1st Arash Ghafouri
*Computer Engineering Department*
*Iran University of Science and Technology*
Tehran, Iran
aghafuri@comp.iust.ac.ir

2nd Hasan Naderi
*Computer Engineering Department*
*Iran University of Science and Technology*
Tehran, Iran
naderi@iust.ac.ir

3rd Mohammad Aghajani
*Physics Department*
*Sharif University of Technology*
Tehran, Iran
m.aghajani@physics.sharif.edu

4th Mahdi Firouzmandi
*Computer Engineering Department*
*Iran University of Science and Technology*
Tehran, Iran
firouzmandi@comp.iust.ac.ir

***Abstract*— Nowadays, one of the main challenges for Question Answering Systems is to answer complex questions using various sources of information. Multi-hop questions are a type of complex questions that require multi-step reasoning to answer. In this article, the IslamicPCQA dataset is introduced. This is the first Persian dataset for answering complex questions based on non-structured information sources and consists of 12,282 question-answer pairs extracted from 9 Islamic encyclopedias. This dataset has been created inspired by the HotpotQA English dataset approach, which was customized to suit the complexities of the Persian language. Answering questions in this dataset requires more than one paragraph and reasoning. The questions are not limited to any prior knowledge base or ontology, and to provide robust reasoning ability, the dataset also includes supporting facts and key sentences. The prepared dataset covers a wide range of Islamic topics and aims to facilitate answering complex Persian questions within this subject matter.**

## I. Introduction

Question Answering (QA) Systems are one of the most important and practical intelligent systems used in the construction of chatbots for social networks and new generation search engines. Question Answering Systems are built using appropriate datasets from unstructured data sources such as encyclopedias and resources, structured data sources such as knowledge graphs, a combination of these two types of resources, and the development of models for using this dataset in order to respond to complex user questions. In other words, these systems are able to receive requests and questions in natural and without semantic limitations to prevent users from wasting time. Then, they provide their answer to the user in the closest and most understandable form possible using advanced methods of information retrieval and argumentation on obtained evidence, in the form of natural language.

Questions asked in the Question Answering System can be simple or complex. Simple questions are usually fact-based, and their answers are typically simple facts or phrases. There are various types of complex questions, of which the most important are comparative (verification or confirmation) questions, multiple-choice questions, bridge questions, and open-ended questions. The complex questions discussed in this article are comparative and bridge questions. In comparative questions, multiple entities or subjects are usually compared with each other. In these questions, a topic can also be verified or rejected, in which case the answer will be yes or no. Bridge questions consist of two or more nested simple questions, for example, if a bridge question can be broken down into two simple questions, the answer to the first simple question will be part of the second simple question. Question Answering systems requires examining at least two different paragraphs to retrieve the answer when facing complex questions or, more precisely, multi-hop questions.

In recent years, despite the presentation of various datasets in the field of Question Answering systems, one of the challenges is the lack of suitable datasets for languages with limited resources such as Persian, and often suitable data resources have been created for English. Obviously, Question Answering systems that rely on limited language datasets such as Persian language Question Answering systems are much less efficient than similar examples developed for English. In Persian language and for Question Answering systems, despite some small efforts, there is still no dataset that can answer multi-hop questions from Persian speakers in advanced search engines or Question Answering systems, which has led to poor performance of Question Answering systems in Persian. Most of the work done on simple Persian language questions, while in reality, most user questions are usually in the form of complex multi-hop questions in these systems.

To address the challenges mentioned in this article, an attempt has been made to improve the performance of these systems in the face of complex multi-hop Islamic questions in Persian language by providing a question answering dataset for such questions. This dataset has been created from non-constructed Islamic data sources and precisely from 9 existing Islamic-Persian encyclopedias.

The organization of the article is as follows. In the next section, the research related to the article is reviewed. The third section fully explains how the dataset was collected and constructed. In the fourth section, some of the modern machine learning techniques and deep neural models are used to validate the IslamicPCQA dataset, and the results obtained are reported. Finally, in the fifth section, some future work is introduced in addition to summarizing the research conducted.

## II. RELATED WORKS

In Persian language, there are five datasets available for question answering systems, which will be discussed below.

PeCoQ[1] dataset is a Persian dataset for knowledge-based question answering, containing 10,000 answers to complex questions from the Persian KB graph. For each question, a SPARQL query is provided. These questions include multi-relational, multi-entity, counting, and constrained questions. In this work, traditional rules and templates were used to generate the dataset.

The PersianQA[2] dataset is a dataset for text comprehension on Persian Wikipedia texts. This dataset contains 9,000 training samples. Some of the questions are unanswerable and some have one or more answers in the background, like the SQuAD2.0[3] dataset. Some questions are impossible and have no answer, which can be useful for systems that know some questions have no answer. This dataset has 900 test data. These questions are collected from historical, religious, geographical, and other thematic categories of articles. Each field in this dataset consists of 7 pairs of questions and one answer, as well as 3 unanswerable questions.

The PQuAD[4] dataset is a dataset designed for machine comprehension, extracted from Persian Wikipedia articles. This dataset contains 80,000 questions along with their answers, of which 25% are unanswerable.

The PersianQuAD[5] dataset is a machine comprehension dataset consisting of approximately 20,000 question-paragraph-answer triplets extracted from Persian Wikipedia articles. This dataset contains difficult questions of various types. The dataset has been evaluated using three models: mBERT[6], ALBERT-FA[7], and ParsBERT[8].

The ParSQuAD[9] dataset is a machine translation-based dataset of the SQuAD 2.0[3] dataset. This dataset is the first large-scale training resource for the Persian language. Many errors in the manual and automatic translation process have been corrected. As a result, this dataset has two versions. Three baseline models, namely BERT[6], ALBERT[7], and MBERT[6], were used to evaluate this dataset. mBERT achieves F1 scores of 56.66% and 52.86% and exact match ratios of 70.84% and 67.73%, respectively, on the test set for version 1 and version 2. This model obtains the best results among the three versions on each version of ParSQuAD.

The ParsSimpleQA[10] dataset is a knowledge graph-based dataset of single-relation simple questions in Persian. The ParsSimpleQA dataset has been semi-automatically created in two stages. In this dataset, first, single-relation question patterns were created, then in the next stage, using templates, entities, and relationships from the Persian base, questions and simple answers were created automatically.

The PerCQA[11] dataset is the first Persian dataset for CQA. It includes questions and answers crawled from one of the most well-known Persian forums. After collecting the data, precise annotation guidelines were provided in a repetitive process, and then question-answer pairs were annotated in the format of SemEvalCQA. PerCQA consists of 989 questions and detailed answers of 21,915 pairs.

In 2014, a text corpus for religious question and answer was developed in Persian, which was limited to religious questions[12]. The raw data used in this corpus includes two text files of Ayatollah Khamenei's verdicts and Ayatollah Makarem Shirazi's treatise provided by the Persian Language Data Refinery. This corpus includes 2051 truth-oriented questions and 2118 non-truth-oriented questions from Ayatollah Khamenei's verdicts and 2456 issues from Ayatollah Makarem Shirazi's treatise. Truth-oriented questions are considered as questions that do not require argumentation to answer. The answers to these types of questions are usually numeric or nominal entities, but non-truth-oriented questions often have longer answers, and due to the need for more argumentation, answering them is more difficult.

There are also multiple datasets for simple and complex question answering in English, but since it is not the subject of this article, describing them is omitted. Table 1 shows all the question answer dataset details available in Persian and English.

## III. DATASET

In this section, the process of creating the dataset is explained.

### *A. Data collection*

The primary goal of this research is to collect the necessary data for generating complex multi-hop questions from non-constructed Islamic resources in Persian. To this end, a collection of Islamic-Persian encyclopedias and famous sources were identified for data collection. Table 2 lists the names and numbers of articles available in these encyclopedias.

### *B. Creating a hyperlink graph*

To formulate a question using the content of Islamic encyclopedias, we need to create a graph-based infrastructure. Since dataset samples or complex questions must be multi-step answerable, it is necessary to construct a graph optimally using the existing links in Islamic encyclopedias, and to ensure that suitable information is provided to the user for asking questions. In the following, a directed graph was created with the help of links in Islamic encyclopedias, where each edge indicates the existence of a link from one page in the encyclopedia (first vertex) to another page in the same encyclopedia or other encyclopedias (second vertex). Various approaches have been proposed and evaluated for constructing this graph, which are explained later.

TABLE I. COMPARING QUESTION ANSWERING DATASETS WITH EACH OTHER

| Dataset | Year | Number of questions | Answer source | Complex question | Language |
|---|---|---|---|---|---|
| WebQuestions[13] | 2014 | 6K | Freebase | × | English |
| WebQuestionsSP[14] | 2016 | 4K | Freebase | ✓ | English |
| LC-QuAD 1.0[15] | 2017 | 5K | DBpedia | ✓ | English |
| LC-QuAD 2.0[16] | 2019 | 30K | DBpedia, Wikidata | ✓ | English |
| FreebaseQA[17] | 2019 | 28K | Freebase | ✓ | English |
| GrailQA[18] | 2021 | 64K | Freebase | ✓ | English |
| VANiLLa[19] | 2018 | 100K | Wikimedia | × | English |
| RuBQ[20] | 2020 | 3K | Wikidata | ✓ | Russian |
| MovieQA[21] | 2015 | 15K | Movies | ✓ | English |
| ParaQA[22] | 2021 | 5K | DBpedia, Wikidata, Yago | ✓ | English |
| ComQA[23] | 2018 | 11K | Wikipedia | ✓ | English |
| VQuAnDa[24] | 2020 | 5K | DBpedia | ✓ | English |
| ConvQuestions[25] | 2019 | 11K | Wikidata | ✓ | English |
| MLQA[26] | 2020 | 12K | Wikipedia | ✓ | Cross-lingual |
| XQuAD[27] | 2020 | 11K | Wikipedia | ✓ | Cross-lingual |
| TriviaQA[28] | 2017 | 950K | Wikipedia, Web | ✓ | English |
| SQuAD 2.0[3] | 2018 | 100K | Wikipedia | ✓ | English |
| CLC-QuAD[29] | 2021 | 28k | Wikidata | ✓ | Chinese |
| CQA-MD[30] | 2019 | 1.5k | CQA | × | Arabic |
| HotpotQA[31] | 2018 | 113K | Wikipedia | ✓ | English |
| PeCoQ[1] | 2020 | 10K | FarsBase | ✓ | Persian |
| PQuAD[4] | 2022 | 80K | Wikipedia | × | Persian |
| PersianQA[2] | 2021 | 10K | Wikipedia | × | Persian |
| PersianQuAD[5] | 2022 | 20K | Wikipedia | × | Persian |
| ParSQuAD[9] | 2021 | 95k | Wikipedia | ✓ | Persian |
| ParsSimpleQA[10] | 2022 | 36k | FarsBase | × | Persian |
| PerCQA[11] | 2022 | 1k | CQA | × | Persian |
| **Our Dataset (IslamicPCQA)** | **2023** | **12k** | **Wikishia, Wikifeqh, Wikiahlolbait, Imamatpedia, Islampedia, Wikihaj, Wikinoor, Wikipasokh, Wikihussain** | ✓ | **Persian** |

Approach 1) Each paragraph as a vertex: In this approach, each paragraph is linked to another related paragraph through the hyperlinks within the text. Only paragraphs with proper length are kept and considered as vertices, while short paragraphs are removed.

Approach 2) The first three paragraphs of each page as a vertex: In this approach, the first three paragraphs of each page that have the necessary characteristics are considered as the graph vertices. The hyperlinks within each paragraph are treated as edges. Typically, each page has at least three paragraphs. Then, one of these paragraphs is paired with a paragraph from another page that is relevant to it, and the resulting two paired paragraphs serve as a sample for generating questions. Although these paragraphs may not necessarily be related, the first two or three paragraphs usually contain general information, making it easier to create high-quality questions.

Approach 3) Each page as a vertex: In this approach, each hyperlink is assigned to the entire page. Therefore, after creating the graph, one of the paragraphs from the first vertex and one of the paragraphs from the second vertex are randomly selected as a related paragraph pair, which is presented as the output. Although this method seems to have very low accuracy, it was still investigated and compared with other methods.

Approach 4) The first six sentences of each page as a vertex: In this approach, the text of the paragraphs is concatenated, and the first six sentences are kept. Then, the hyperlinks within these six sentences are considered as the edges. This technique ensures that the paragraphs have a consistent length, making it easier to formulate questions. Although this method generates fewer related paragraph pairs, it is still suitable because the number of results is approximately double the number of pages available. Despite the fact that the selected text is more general and easier to question, the number of related paragraph pairs generated is lower.

TABLE II. ISLAMIC ENCYCLOPEDIAS HAVE BEEN CRAWLED AS A SOURCE OF DATA.

| The name of the encyclopedia | Number of articles (Persian language) |
|---|---|
| wikishia.net | 7.000 |
| wikifeqh.ir | 71.000 |
| wiki.ahlolbait.com | 14.200 |
| imamatpedia.com | 39.500 |
| islampedia.ir | 2.000 |
| wikihaj.com | 1.500 |
| wikihussain.com | 15.000 |
| wikipasokh.com | 2.000 |
| wikinoor.ir | 3.500 |
| Overall | 155.700 |

The initial analysis of encyclopedias showed that the third approach for constructing hyperlinked graphs was not suitable. Therefore, three other approaches were used for implementation, and based on a general analysis of encyclopedia data, these approaches led to the construction of graphs in the most optimal state. To summarize and select a unified approach for graph construction, the output of these three approaches was examined on 500 sample data, and finally the fourth approach was selected for graph construction.

The fourth approach had numerous features and advantages compared to the others, which are mentioned below:

- All produced text pairs have a fixed number of sentences (6 sentences), and therefore it can be confidently stated that the number of sentences provided to the user is sufficient.
- The separated text, or the first six sentences, has a high level of generality and information. Observations indicate the fact that the beginning sentences of each encyclopedia page naturally have general and suitable information, and as time goes by, more detailed information is provided that may not necessarily have much reliable or significant relation to the hyperlinks present in them. As a result, the easier and higher-quality question design is put forward on the principles.
- Each text pair is "definitely" related to each other based on a "word or part of the text"; the word that appeared in blue as a hyperlink in the page's text is searched, and if it exists, a link is established between the two paragraphs.

After finding the optimal method for producing text pairs, the production of paragraph pairs from each of the available encyclopedias was carried out. Finally, 637,700 paragraph pairs were produced from various encyclopedias, and the distribution of produced paragraph pairs is visible in Table 3.

TABLE III. THE DISTRIBUTION OF PRODUCED PARAGRAPH PAIRS FOR EACH ENCYCLOPEDIA AND ITS CORRESPONDING PROBABILITY DISTRIBUTION WAS ANALYZED.

| **Islamic encyclopedia** | **Even number of production paragraphs** | **Probability distribution (prioritization)** |
|---|---|---|
| wikishia.net | 49.000 | 0.3 |
| wikifeqh.ir | 245.000 | 0.22 |
| wiki.ahlolbait.com | 260.000 | 0.13 |
| imamatpedia.com | 42.500 | 0.12 |
| islampedia.ir | 30.000 | 0.1 |
| wikihaj.com | 4.500 | 0.05 |
| wikihussain.com | 6.000 | 0.04 |
| wikipasokh.com | 500 | 0.03 |
| wikinoor.ir | 200 | 0.01 |
| Overall | 637.700 | 1 |

In normal circumstances, large encyclopedias allocate more samples to themselves due to their high number of pages. However, since the content of these encyclopedias does not necessarily have the necessary richness, generating data samples faces challenges. To solve this challenge and generate data samples using encyclopedias that have a much more suitable status in terms of criteria such as topic diversity, complexity level, and level of content correlation in the text, a probability distribution was considered for each of the encyclopedias. Then, attempts were made to display more samples from the encyclopedia with richer content. After producing appropriate paragraph pairs, a user interface was designed so that people could design complex questions with the given paragraph pairs and add to the number of records in the dataset. Figure 1 shows an overview of the process of creating the IslamicPCQA dataset.

## IV. DATASET ANALYSIS

As mentioned in previous sections, the IslamicPCQA dataset contains 12,282 complex question-answer data samples. Among these data samples, 800 were randomly selected and their questions and answers were analyzed and studied. The results of analyzing the question-answer data samples are explained below.

### *A. Types of Reasoning*

This dataset includes various types of multi-step reasoning to reach the correct answer, which will be discussed below and, in addition to categorization, the percentage of each category is also presented. There are five types of reasoning to answer user questions in the IslamicPCQA dataset, along with the percentage frequency and an example in Table 4.

Based on the structure of the dataset, almost all data samples require at least one guidance sentence from each paragraph. In reasoning type 1, with a frequency percentage of 43.2%, to reach the answer, the first step is to find an intermediary entity between two paragraphs. Then, the second step of reasoning begins and the final answer is extracted. As seen in the example for this type, "Shaban" is the intermediary entity between two paragraphs, and by taking the second reasoning step through it, the final answer, which is "Prophet Muhammad (PBUH)", is reached. In general, in type 1 questions, to take the second reasoning step and answer the question directly (like the example mentioned) or indirectly (i.e., through the relevant feature of this intermediary entity), an intermediary entity between two paragraphs is needed. In reasoning type 2, with a frequency percentage of 2.3%, as shown in the example in Table 4, a comparison must be made between two paragraphs and by comparing the two entities or the feature related to these two entities, which can be quantitative or qualitative, the answer is obtained. In reasoning type 3, with a frequency percentage of 6.5%, several features need to be examined to reach the final answer. The approach to answering these types of questions is to extract all relevant features to the question and then the final answer is obtained by establishing a correct (or shared) relationship between these features. In the example related to this type, it is necessary to first extract "all the worldly consequences of occultation" as relevant features to the question. Then by examining the second paragraph, the desired consequence that the question is concerned with and has in common with it is found. In reasoning type 4, with a frequency percentage of 11%, reasoning through the intermediary entity is required, and the relevant feature of the main entity is found, and the final answer is extracted. As shown in the example for this type, the location of "the battle of Imam Ali (AS) with Amr ibn Abd al-Wud" as the feature of this entity is obtained through "Khandaq War" which is the intermediary entity. Finally, in reasoning type 5, which has a frequency percentage of 37%, there are a set of questions that require more than 2 guidance sentences from the golden paragraphs to move towards the answer to the question. Most of these questions require more reasoning steps and are therefore more complex, which shows the high quality of the dataset.

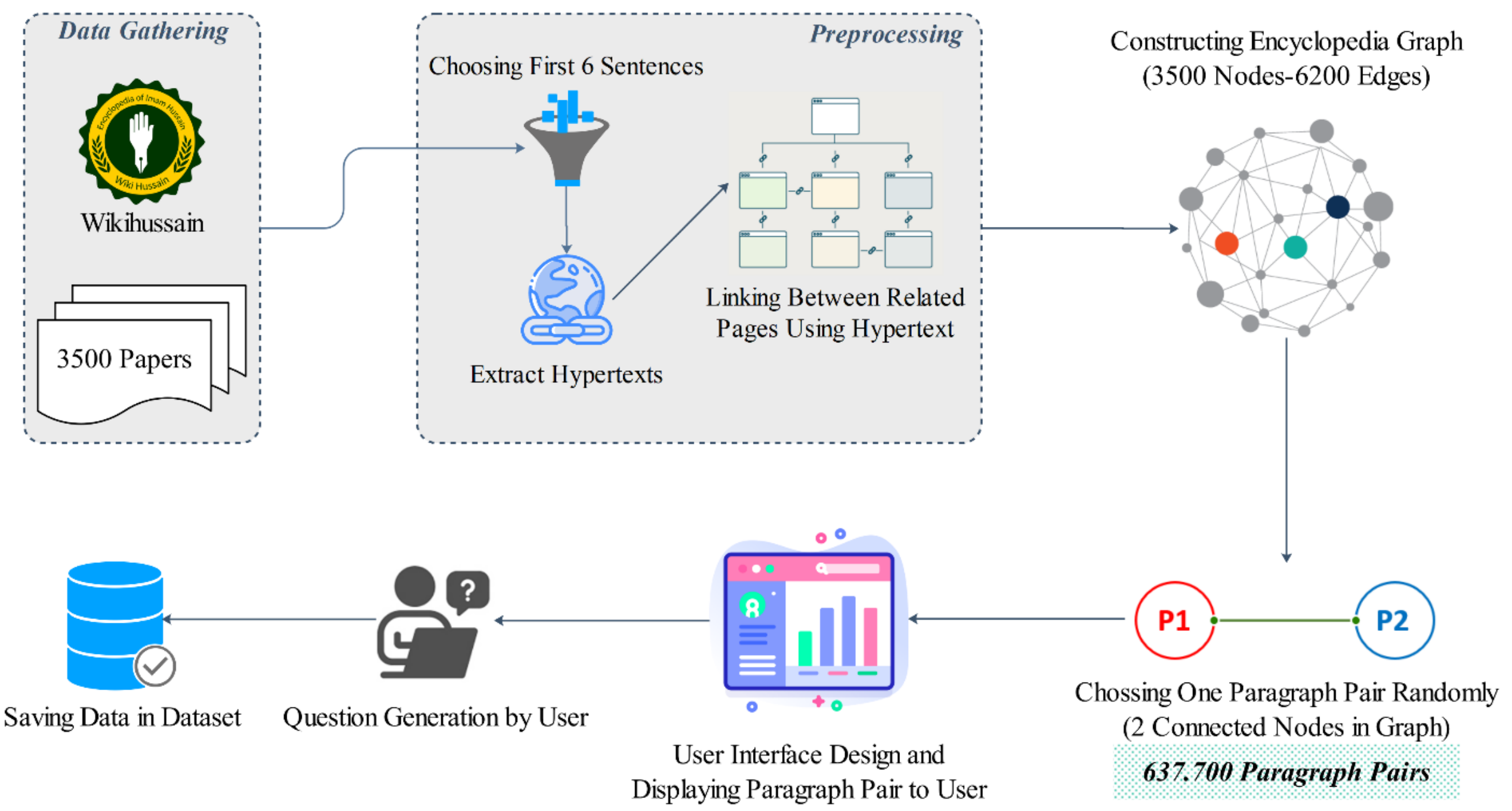

Fig. 1. The overall picture of the process of creating the IslamicPCQA dataset.

TABLE IV. TYPES OF MULTI-HOP REASONING IN THE ISLAMICPCQA DATASET.

| Type of arguments | Frequency | Example |
|---|---|---|
| Type 1: Reaching the answer by using reasoning through an intermediary entity. | 43.2 | **پاراگراف اول**: شعبان یا شَعبان المُعَظّم هشتمین ماه تقویم هجری قمری است ... که در این ماه مناسبت‌های مذهبی مختلفی وجود دارد از جمله آنها میلاد امام حسین(ع)، حضرت عباس(ع)، امام سجاد(ع)، و حضرت علی‌اکبر و امام مهدی(عج) است که به اعیاد شعبانیه شناخته می‌شوند.<br>**First paragraph**: Shaban or Sha'ban al-Muazzam is the eighth month of the lunar hijri calendar... Various religious occasions are observed during this month, including the birthdays of Imam Hussein, Hazrat Abbas, Imam Sajjad, Hazrat Ali Akbar, and Imam Mahdi, which are known as the Shabanian festivals.<br>**پاراگراف دوم**: شعبان از ریشه شعب است و از آنجا که در این ماه روزی مؤمنان و حسنه در آن منشعب و بسیار می‌گردد، این ماه شعبان نام گرفت ... براساس روایات، ماه شعبان، ماه پیامبر اکرم(ص) است.<br>**Second paragraph**: Shaban derives from the root word Shu'ub, meaning branching out or spreading. It is named Shaban because during this month, believers perform many good deeds and righteousness spreads. According to traditions, Shaban is the month of Prophet Muhammad (PBUH).<br>**سوال**: ماه میلاد امام حسین(ع) به ماه کدام پیامبر معروف است؟<br>**Question**: In which month is the birth of Imam Hussein (AS) and which famous prophet does this month belong to? |
| Type 2: Reaching the answer by comparing two entities. | 2.3 | **پاراگراف اول:** شعبان یا شَعبان المُعَظّم هشتمین ماه تقویم هجری قمری است که بنا به احادیث، فضیلت بسیاری دارد و در آن به روزه‌داری، صدقه دادن، صلوات فرستادن و خواندن دعاهای خاص مانند صلوات شعبانیه و مناجات شعبانیه سفارش شده است<br>**First paragraph**: Sha'ban, also known as Sha'ban al-Muazzam, is the eighth month of the lunar Hijri calendar. According to Hadiths, it has many virtues, and in it, fasting, charity, sending blessings, and reciting specific prayers such as Sha'ban supplication and invocation are recommended.<br>**پاراگراف دوم:** رَجَب یا رَجَب المُرَجّب هفتمین ماه از ماه‌های قمری و نخستین ماه از ماه‌های حرام است.<br>**Second paragraph**: Rajab, also known as Rajab al-Murajjab, is the seventh month of the lunar Hijri calendar and the first month of the sacred months.<br>سوال: آیا ماه رجب زودتر از ماه شعبان است؟(بله)<br>**Question**: Is Rajab month earlier than Sha'ban month? (Yes) |
| Type 3: Finding the answer entity by examining multiple features. | 6.5 | **پاراگراف اول:** پیامدهای غیبت به دو بخش دنیوی و اخروی تقسیم شده است ... ایجاد بی‌اعتمادی نسبت به غیبت کننده و ایجاد کینه و دشمنی در جامعه، برخی از پیامدهای دنیوی غیبت است.<br>**First paragraph**: The consequences of backbite have been divided into two worldly and other-worldly sections ... creating distrust towards the backbite and creating hate and enmity in society are some of the worldly consequences of backbite.<br>**پاراگراف دوم:** کینه یکی از رذائل اخلاقی و حالتی روحی است که فرد در آن حالت، دشمنیِ دیگری را در دل پنهان کرده و منتظر فرصت مناسب است تا آن را ابراز کند ... کینه موجب گناهان دیگری مانند حسادت، غیبت، تهمت، سرزنش، توبیخ و تحقیر دیگران نیز می‌شود.<br>**Second paragraph**: Hatred is one of the moral vices and a mental state in which an individual hides |
| | | animosity towards someone and waits for the right opportunity to express it... Hatred leads to other sins such as jealousy, backbiting, slander, reproach, admonishment, and belittling of others.<br>**سوال**: کدام پیامد دنیوی غیبت است که فرد در آن حالت، دشمنیِ دیگری را در دل پنهان کرده است؟<br>**Question:** What is the worldly consequence of backbiting in which an individual hides animosity towards someone in their heart? |
| Type 4: Reasoning to find the feature of an entity through an intermediary entity. | 11 | **پاراگراف اول:** حضرت علی(ع) در همه غزوات پیامبر اسلام(ص) به جز جنگ تبوک شرکت داشت ... در غزوه خندق حضرت علی(ع) با کشتن عمر بن عبدود جنگ را خاتمه داد و در غزوه خیبر، با کندن درِ بزرگ قلعه خیبر، جنگ را به سرانجام رساند.<br>**First paragraph**: Imam Ali (AS) participated in all the battles of Prophet Muhammad (PBUH) except for the Battle of Tabuk. In the Battle of Khandaq, Imam Ali (AS) put an end to the war by killing Amr ibn Abd Wudd. In the Battle of Khaybar, Imam Ali (AS) conquered the great fortress of Khaybar by breaking down its gate and bringing the battle to an end.<br>**پاراگراف دوم:** غزوه خندق یا غزوه احزاب از غزوات پیامبر که در سال پنجم هجری قمری رخ داد ... برای مقابله با مشرکان به پیشنهاد سلمان فارسی، مسلمانان در اطراف مدینه خندق حفر کردند.<br>**Second paragraph:** The Battle of the Trench or the Battle of the Confederates was one of the Prophet's battles that took place in the fifth year of the Hijri lunar calendar. The Muslims dug trenches around Medina to confront the polytheists at the suggestion of Salman Farsi.<br>**سوال**: حضرت علی(ع) در کجا با عمربن عبدود جنگید و او را شکست داد؟<br>**Question:** Where did Imam Ali (AS) fight with Amr bin Abdoud and defeat him? |
| Type 5: Other reasoning methods that require more than two guidance sentences. | 37 | **پاراگراف اول:** سال ۱۴۹ هجری قمری یکصد و چهل و نهمین سال از سال‌شمار هجری قمری و مصادف با امامت امام کاظم(ع) و زندگانی امام رضا(ع) است ... در این سال، منصور عباسی از سلسله عباسیان بر عراق، و عبدالرحمن الداخل از دودمان امویان، بر اسپانیا حکومت می‌کردند.<br>**First paragraph:** In the year 149 AH, which corresponds with the Imamate of Imam Kazem (AS) and the life of Imam Reza (AS), Mansour Abbasi from the Abbasid dynasty ruled over Iraq, and Abdulrahman al-Dakhil from the Umayyad dynasty ruled over Spain.<br>**پاراگراف دوم:** ابوجعفر عبدالله منصور مشهور به منصور دَوانیقی (حکومت:۱۳۶-۱۵۸ق)، دومین خلیفه عباسی و از نوادگان عباس بن عبدالمطلب ... او هم‌عصر امام صادق(ع) بود و بنابر نقل ابن شهر آشوب، ایشان را به شهادت رساند.<br>**Second paragraph:** Abu Ja'far Abdullah Mansur, known as Mansur al-Dawaniqi, ruled from 136-158 AH, as the second caliph of the Abbasid dynasty and one of the descendants of Abbas ibn Abdul Muttalib. He was contemporaneous with Imam Sadegh (AS) and according to Ibn Shahr Ashub, he caused his martyrdom.<br>**سوال**: فردی که سال ۱۴۹ قمری در سلسله عباسیان بر عراق حکومت می کرد هم عصر کدام امام بوده است؟<br>**Question:** The person who ruled over Iraq in the Abbasid dynasty in the year 149 AH was contemporary with which Imam? |

## *B. Analysis of questions*

This section examines the statistical analysis of the dataset questions.

**The number of questions with various interrogative words used**: The IslamicPCQA dataset includes a wide range of types of questions that start with various interrogative words such as "why", "how", "what is", and so on. In Table 5 and Figure 2, a detailed categorization of the number of questions is displayed based on the type of interrogative word used in them.

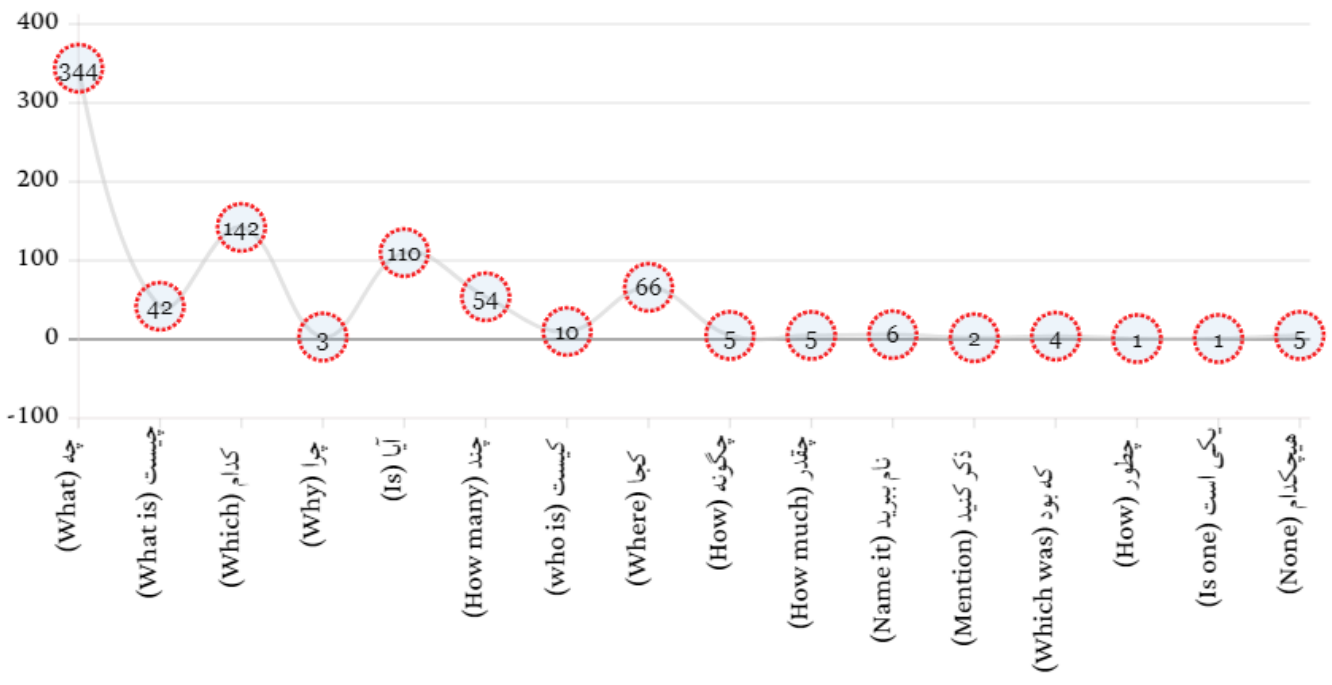

Fig. 2. Categorizing questions based on interrogative words used.

TABLE V. DETAILS OF CATEGORIZING QUESTIONS BASED ON INTERROGATIVE WORDS USED.

| Question keyword type | Number | Frequency |
|---|---|---|
| (what) چه | 344 | 43 |
| (what is) چیست | 42 | 5.25 |
| (which) کدام | 142 | 17.75 |
| (why) چرا | 3 | 0.375 |
| (is) آیا | 110 | 13.75 |
| (how many) چند | 54 | 6.75 |
| (who is) کیست | 10 | 1.25 |
| (where) کجا | 66 | 8.25 |
| (how) چگونه | 5 | 0.625 |
| (how much) چقدر | 5 | 0.625 |
| (name it) نام ببرید | 6 | 0.75 |
| (mention) ذکر کنید | 2 | 0.25 |
| (which was) که بود | 4 | 0.5 |
| (how) چطور | 1 | 0.125 |
| (is one) یکی است | 1 | 0.125 |
| (none) هیچکدام | 5 | 0.625 |

**The number of questions for different types of arguments**: In previous sections, 5 types of arguments were introduced for dataset questions. In the following, in Table 6 and Figure 3, the percentage and number of questions based on the type of argument they support can be observed.

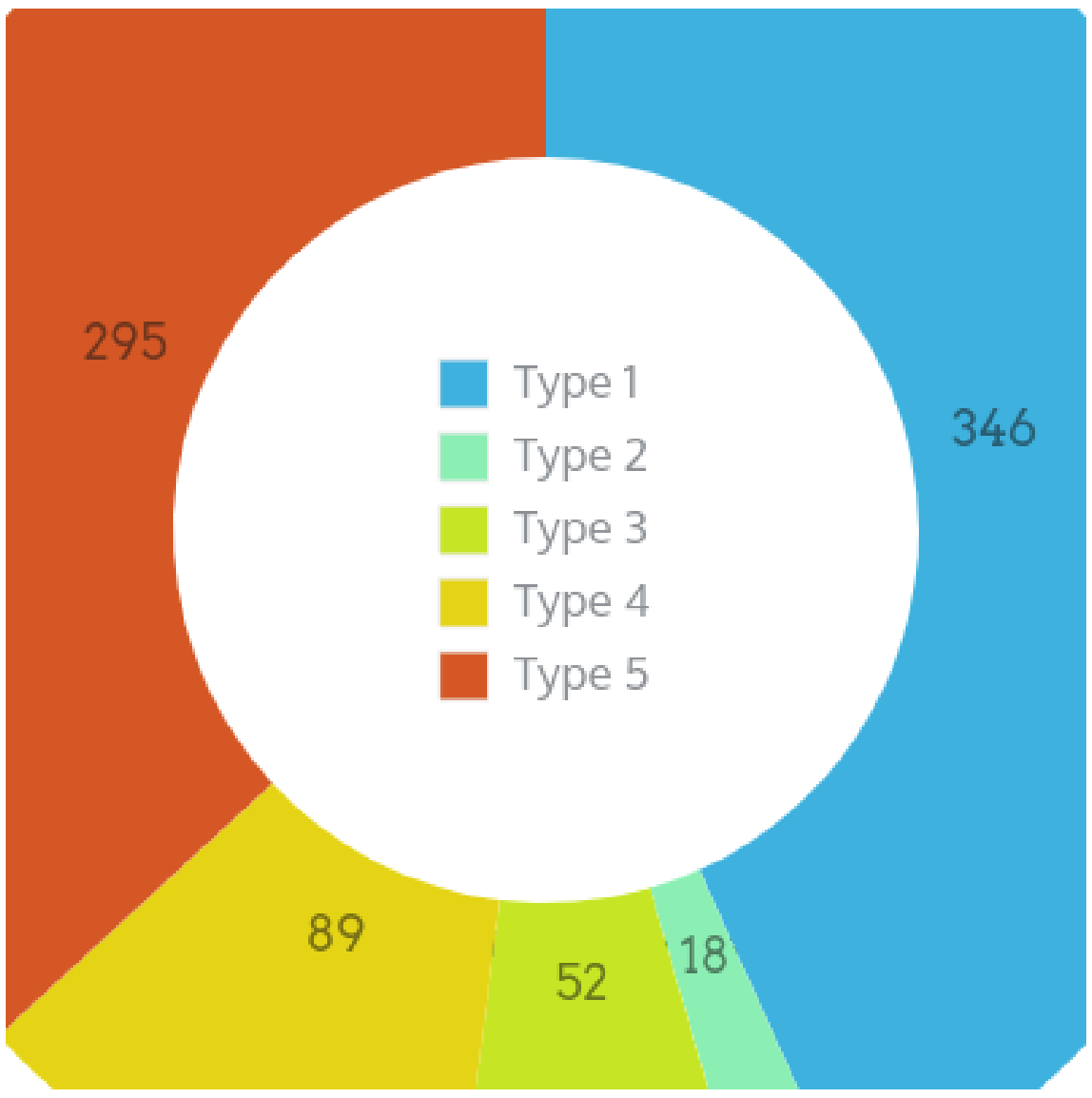

Fig. 3. Categorizing of questions based on 5 different types.

TABLE VI. THE DETAILS OF QUESTION CATEGORIZATION BASED ON 5 DIFFERENT TYPES.

| Question type | Number | Frequency |
|---|---|---|
| Type 1 | 346 | 43.25 |
| Type 2 | 18 | 2.25 |
| Type 3 | 52 | 6.5 |
| Type 4 | 89 | 11.125 |
| Type 5 | 295 | 36.875 |

**Number of characters in questions**: The IslamicPCQA dataset includes various long and short questions. These questions have been examined based on the number of characters, and the results are displayed in figure 4.

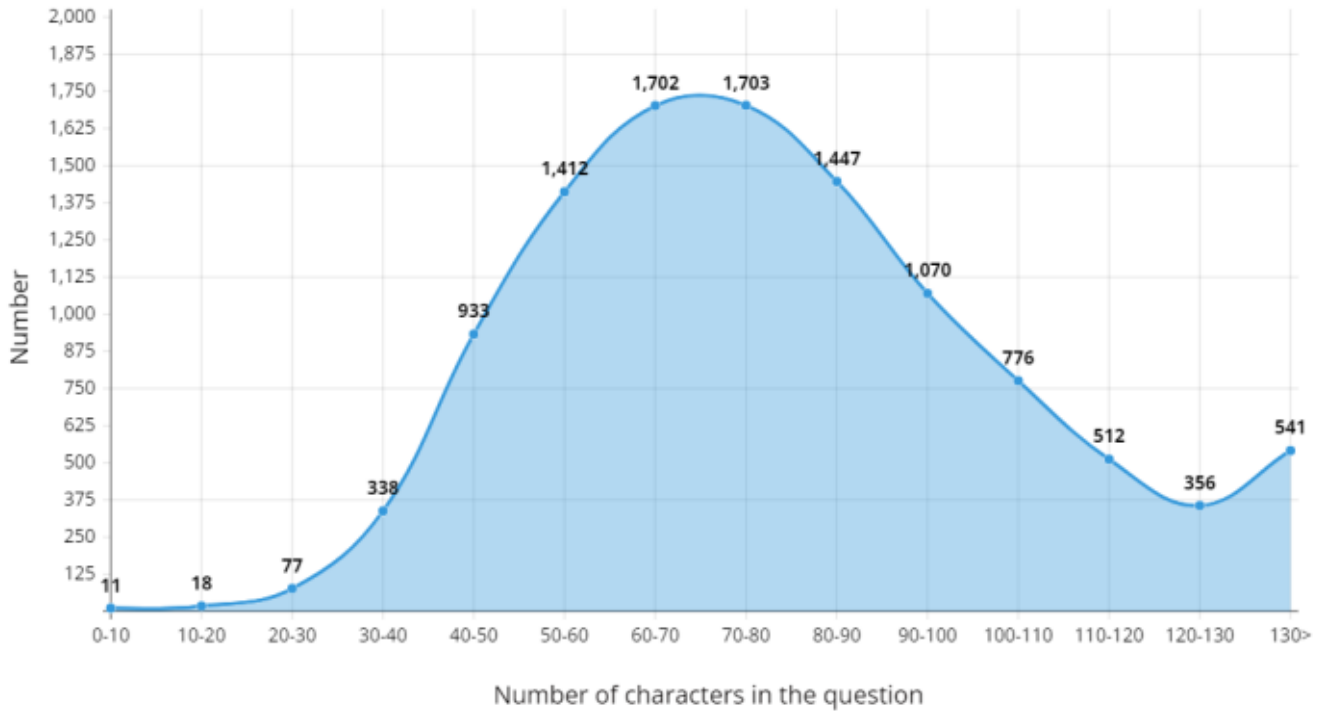

Fig. 4. Question categorization based on the number of characters in the question.

## C. *Analysis of answers*

The IslamicPCQA dataset consists of a wide range of question types that start with various interrogative words such as "why", "how", "what" and others. Table 5 and Figure 2 show a precise categorization of the number of questions based on the type of interrogative word used. The questions and subsequent answers of the IslamicPCQA dataset are grouped into twelve general categories in terms of subject diversity. These twelve categories include person, group, location, date, number, artwork, yes/no, event, proper noun, common noun, and phrase. Table 6 and Figure 5 show the frequency of different types of answers. As can be observed, the most common type of answer is related to "person" and the least common type of answer is related to "event".

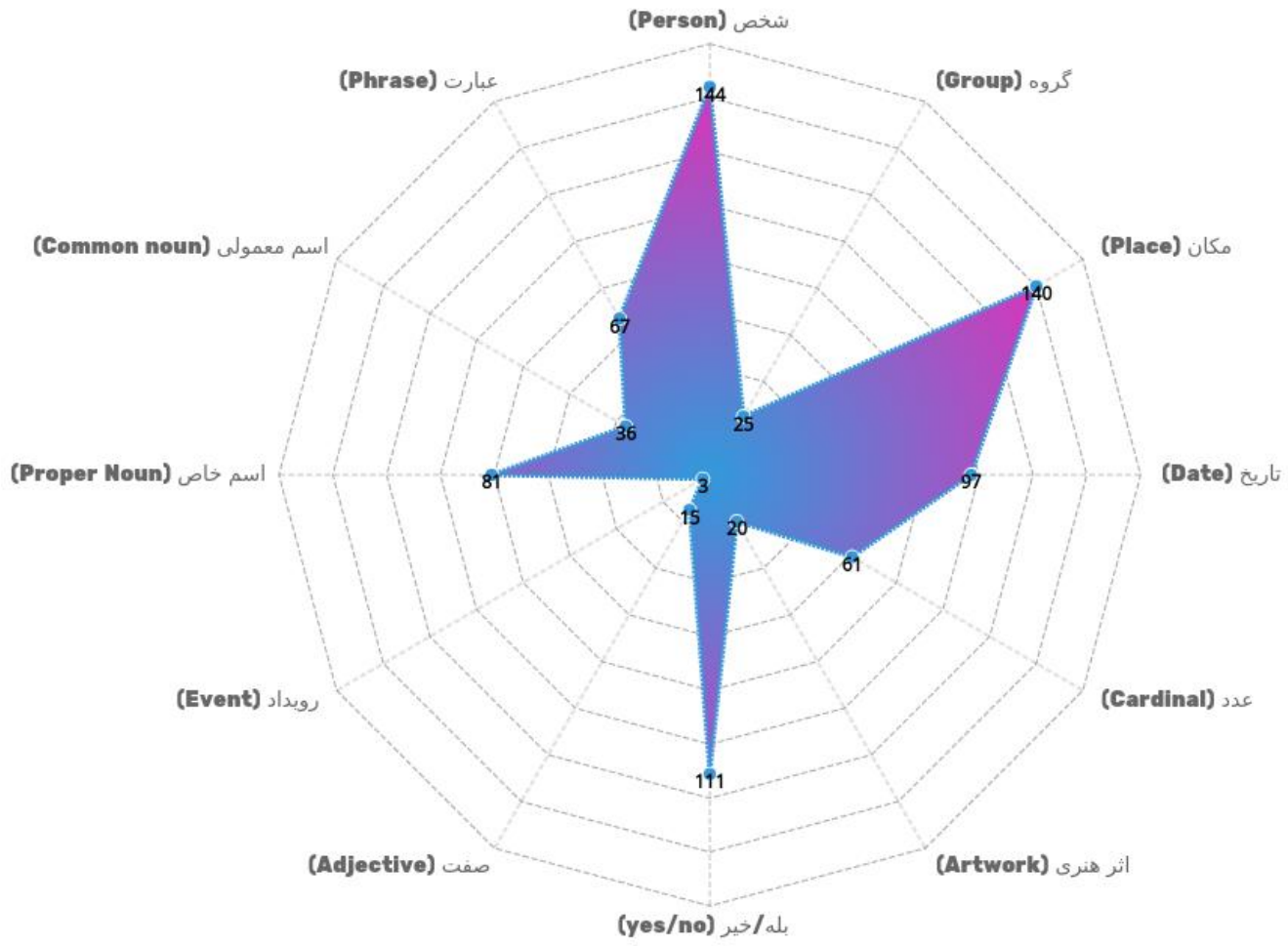

Fig. 5. Categorizing of answers to questions based on 12 different types.

TABLE VII. THE DETAILS OF CATEGORIZING THE ANSWERS OF QUESTIONS INTO 12 DIFFERENT TYPES ARE AS FOLLOWS.

| Type of answer | Number | Frequency |
|---|---|---|
| (Person) شخص | 144 | 18 |
| (Group) گروه | 25 | 3.125 |
| (Place) مکان | 140 | 17.5 |
| (Date) تاریخ | 97 | 12.125 |
| (Cardinal) عدد | 61 | 7.625 |
| (Artwork) اثر هنری | 20 | 2.5 |
| (Yes/No) بله/خیر | 111 | 13.875 |
| (Adjective) صفت | 15 | 1.875 |
| (Event) رویداد | 3 | 0.375 |
| (Proper Noun) اسم خاص | 81 | 10.125 |
| (Common Noun) اسم معمولی | 36 | 4.5 |
| (Phrase) عبارت | 67 | 8.375 |

## D. *Dataset benchmarking method*

Using the process described in Section 3, approximately 12,280 data samples were generated by different users. From the dataset, randomly selected 9,000 data samples were allocated for the train section, 1,641 data samples for the dev section, and 1,641 data samples for the test section. Different models were considered for training and testing purposes. The dataset includes Islamic information.

TABLE VIII. SPLIT THE SAMPLE DATA.

| Number of data samples | Application | File name |
|---|---|---|
| 9.000 | training | train |
| 1.641 | dev | dev_distractor |
| 1.641 | test | test_fullwiki |
| 12.282 | Overall | |

Two types of adjustments have been considered for evaluating neural network models using this dataset, which will be explained in the following.

**Setting the first type or distractor**: In the first setting, also known as a distractor, there are 2 golden paragraphs that the question is generated from, and 8 other paragraphs that are very similar to them, and the models need to be able to identify the two golden paragraphs and find the answer to the question while ignoring the distractors. To create this distractor setting, using the retrieval section of the DrQA project which is built by the TF-IDF+Bigram method, the first 8 related paragraphs to the question are extracted. Then, 2 golden paragraphs[1] are added to these 8 related paragraphs and finally, a total of 10 paragraphs related to the question are saved in the dataset. This process provides a basis for the model to learn to find and select the correct guide sentences[2], which is a feature of this type of data sample generation.

**Setting the second type or fullwiki**: In the creation of another configuration named fullwiki, the system must extract 10 relevant paragraphs through retrieval and store them in the data set, similar to the previous configuration. Then, similar to the direct configuration, the two most relevant paragraphs are identified and the answer to the question is determined. To create the second configuration or fullwiki, 10 relevant paragraphs are extracted by the system using the TF-IDF+Bigram method, just like the direct configuration, and stored in the data set. In this configuration and in retrieving its related paragraphs, it is not necessarily certain that the golden paragraphs are present, and the presence or absence of golden paragraphs depends on the accuracy of paragraph retrieval. The feature of this type of configuration is the production of a sample data file and complete preparation of the model for finding the answer and guidance sentences; a serious benchmark for what happens in practice. Generally, the trained model's ability to find the correct answer when there are no golden paragraphs (or one of them) among the 10 related paragraphs found by the retrieval system is highly influenced. Therefore, the model must be trained with a large and diverse set of data to have the power to find the necessary features to reach the answer in the absence of golden paragraphs among the 10 relevant paragraphs.

## V. TESTING AND EVALUATING DATASETS

To evaluate the IslamicPCQA dataset, four pre-trained models, XLM-RoBERTa, ParsBert, BERT multilingual, and mT5, were selected. The results of each of these models were presented separately based on the two criteria F1 Score and Exact Match on the mentioned dataset. Figure 6 provides an overview of how language models are trained on the IslamicPCQA dataset.

TABLE IX. MODEL FINE-TUNING CONDITIONS WITH THE ISLAMICPCQA DATASET

| Condition | Value |
|---|---|
| learning_rate | 3e-5 |
| per_device_train_batch_size | 16 |
| per_device_eval_batch_size | 16 |
| num_train_epochs | 3 |
| weight_decay | 0.0001 |

***1) XLM-RoBERTa:*** The XML-RoBERTa[32] model has been trained on 2.5 terabytes of data collected from 100 different languages. In this experiment, the dataset was fine-tuned on this model with 3 epochs. The results obtained based on F1 and exact match metrics on the Distractor tuning set are 80.44 and 67.22, respectively. The results obtained based on the mentioned evaluation metrics on the Fullwiki tuning set are 72.86 and 56.42, respectively.

***2) ParsBert:*** In this experiment, the ParsBERT[8] model was used, which is a BERT-based monolingual language model. The model was trained on a large Persian dataset with various writing styles and topics, including scientific articles, novels, news, and more, consisting of 3.9 million documents, 73 million sentences, and 1.3 billion words. The dataset was fine-tuned on this model with 3 epochs. The results obtained based on F1 and exact match metrics on the Distractor tuning set are 75.84 and 63.86, respectively. The results obtained based on the mentioned evaluation metrics on the Fullwiki tuning set are 63.41 and 47.59, respectively.

***3) BERT multilingual:*** A multilingual BERT[6] language model was trained on Wikipedia articles from 102 different languages. The model was fine-tuned on the IslamicPCQA training dataset for 3 epochs. In this experiment, the results based on F1 and exact match metrics on the Distractor tuning dataset are 72.37 and 57.86, respectively. Also, the results on the Fullwiki tuning dataset for the mentioned evaluation metrics are 64.55 and 47.95, respectively.

***4) mT5:*** The mT5[33] model is a pre-trained multilingual text-to-text Transformer. The model was trained on datasets from 101 different languages. The model was fine-tuned on the IslamicPCQA training dataset as the last experiment. The results based on F1 and exact match metrics on the Distractor tuning dataset are 61.28 and 21.61, respectively. Also, the results on the Fullwiki tuning dataset for the mentioned evaluation metrics are 48.59 and 16.11, respectively.

The results of fine-tuning the models on the dataset created in table number 8 are visible. Based on the experiments, the XLM-Roberta-Large model has the best results, and mT5 has the worst results in both tuning scenarios.

[1] The term "two golden paragraphs" refers to two main paragraphs that questions are designed based on.
[2] supporting facts

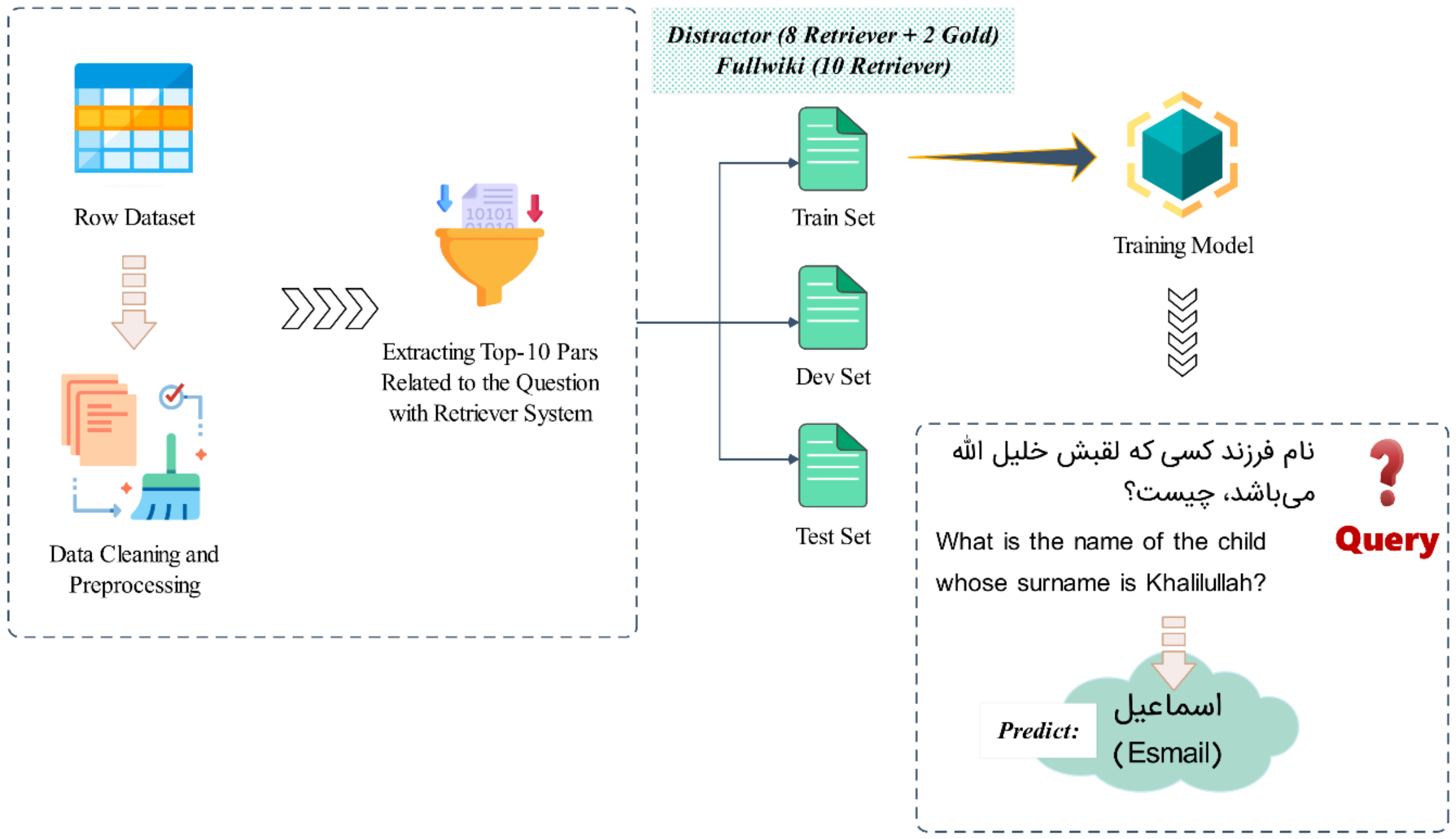

Fig. 6. The general overview of fine-tuning models for evaluating the IslamicPCQA dataset is as follows.

In the mT5 model, the results of the Exact Match metric are relatively lower than other models due to its text-to-text structure. In this structure, a text is given as input, and the output is also a text. Therefore, the generated texts as answers are close to the original answer in terms of structure, but not the same. As a result, the Exact Match metric score is lower. Because the Exact Match metric requires the predicted output of the model to be the same as the output of the Test dataset.

TABLE X. COMPARING THE RESULTS OF DIFFERENT TRAINED MODELS ON THE ISLAMICPCQA DATASET.

| Dataset | Distractor | | Fullwiki | |
|---|---|---|---|---|
| **Model** | **F1** | **EM** | **F1** | **EM** |
| **Our XLM-Roberta-Large** | 80.44 | 67.33 | 72.86 | 56.42 |
| **Our ParsBERT** | 75.84 | 63.86 | 63.41 | 47.59 |
| **Our Multilingual BERT** | 72.37 | 57.86 | 64.55 | 47.95 |
| **Our mT5** | 61.28 | 21.61 | 48.59 | 16.11 |

In Figure 7, the performance results of the baseline models on the test dataset can be compared based on the Exact Match criterion.

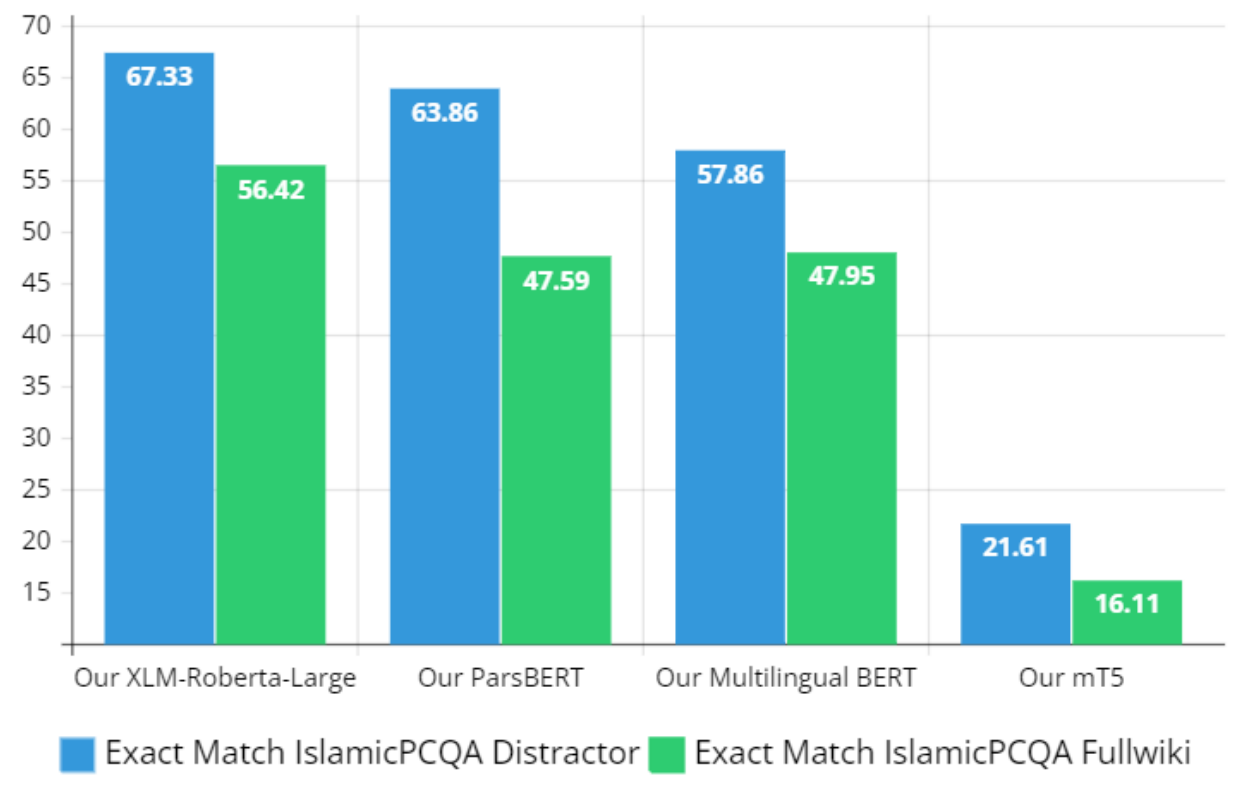

Fig. 7. Comparison of the evaluation results of 4 trained models on the Distractor and Fullwiki datasets based on Exact Match criterion.

In Figure 8, the comparison of performance results of baseline models on the test dataset based on the F1 Score criterion can be seen.

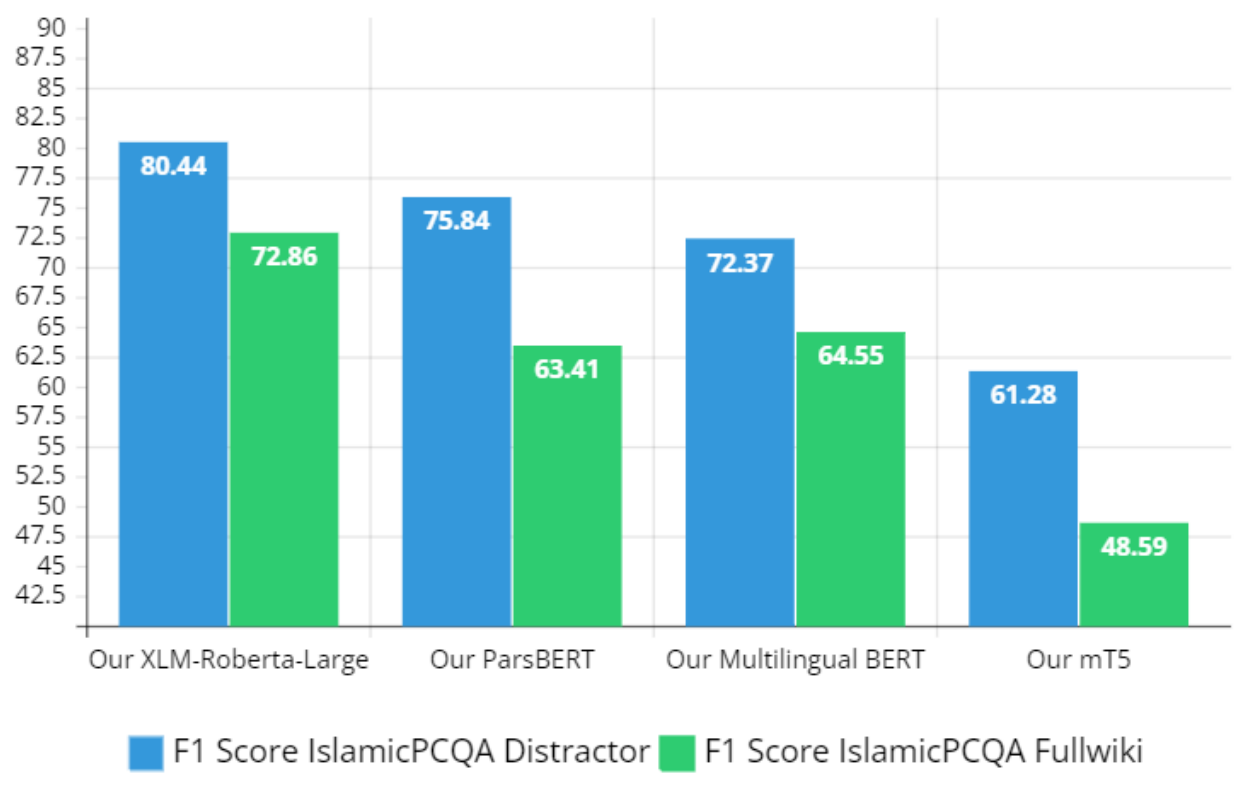

Fig. 8. The comparison of the evaluation results of 5 trained models on the Distractor and Fullwiki datasets based on the F1 Score criterion.

A summary of the evaluation results of the trained models on the IslamicPCQA Distractor dataset is shown in Figure 9.

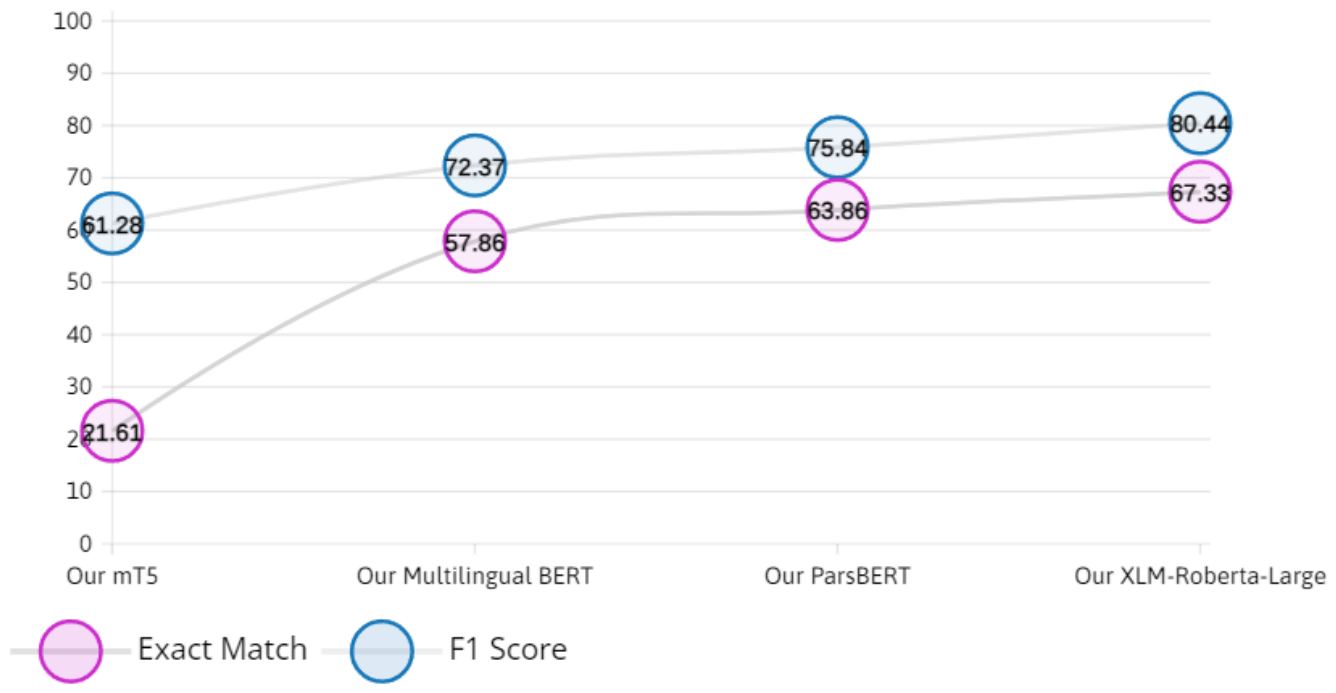

Fig. 9. Comparison of evaluation results on the IslamicPCQA Distractor dataset for different models is shown.

It is a positive and important point that for some questions that fall into the "acceptable" and "incorrect (but partially correct)" categories, examining the extracted answer confirms that the absence of two Golden Paragraphs side by side and the presence of only one of them, caused the system to only take one reasoning step and not be able to take another step to reach the answer. Therefore, although the performance can be improved, taking into account these points and conditions gives hope for the positive performance of the trained model. It also seems that if the questions are simple or one-step, the system is powerful enough to find the correct answers with high accuracy.

## VI. Conclusion

In this research, a dataset was created to facilitate answering complex Persian and Islamic questions that many audiences have today. This dataset was created by extracting from the largest and most famous Persian language and Islamic encyclopedias and with the help of many linguistic experts. This dataset can answer various complex multi-hop questions such as bridge and comparison and supports various types of arguments. To evaluate the efficiency of the created dataset, it was tested on F1 and Exact Match criteria using the latest neural transformer-based models and achieved acceptable results. In future work, the mentioned dataset can be expanded by using other Persian encyclopedias such as Wikipedia, which are open domain and have more records, and for the public domain, including non-Islamic domains. On the other hand, better results can be achieved on this dataset by using innovative methods and developing newer models.

## References

[1] R. Etezadi and M. Shamsfard, “PeCoQ: A Dataset for Persian Complex Question Answering over Knowledge Graph,” in 2020 11th International Conference on Information and Knowledge Technology (IKT), Dec. 2020, pp. 102–106. doi: 10.1109/IKT51791.2020.9345610.

[2] S. Ayoubi, “https://github.com/sajjjadayobi/PersianQA.”

[3] P. Rajpurkar, R. Jia, and P. Liang, “Know What You Don’t Know: Unanswerable Questions for SQuAD.” 2018.

[4] K. Darvishi, N. Shahbodagh, Z. Abbasiantaeb, and S. Momtazi, “PQuAD: A Persian Question Answering Dataset,” Feb. 2022.

[5] A. Kazemi, J. Mozafari, and M. A. Nematbakhsh, “PersianQuAD: The Native Question Answering Dataset for the Persian Language,” IEEE Access, vol. 10, pp. 26045–26057, 2022, doi: 10.1109/ACCESS.2022.3157289.

[6] J. Devlin, M. W. Chang, K. Lee, and K. Toutanova, “BERT: Pre-training of deep bidirectional transformers for language understanding,” in NAACL HLT 2019 - 2019 Conference of the North American Chapter of the Association for Computational Linguistics: Human Language Technologies - Proceedings of the Conference, 2019, vol. 1, pp. 4171–4186. [Online]. Available: https://github.com/tensorflow/tensor2tensor

[7] Z. Lan, M. Chen, S. Goodman, K. Gimpel, P. Sharma, and R. Soricut, “ALBERT: A Lite BERT for Self-supervised Learning of Language Representations.” 2020.

[8] M. Farahani, M. Gharachorloo, M. Farahani, and M. Manthouri, “ParsBERT: Transformer-based Model for Persian Language Understanding,” Neural Process. Lett., vol. 53, no. 6, pp. 3831–3847, 2021, doi: 10.1007/s11063-021-10528-4.

[9] N. Abadani, J. Mozafari, A. Fatemi, M. A. Nematbakhsh, and A. Kazemi, “ParSQuAD: Machine Translated SQuAD dataset for Persian Question Answering,” in 2021 7th International Conference on Web Research (ICWR), 2021, pp. 163–168. doi: 10.1109/ICWR51868.2021.9443126.

[10] H. Babaei Giglou, N. Beyranvand, R. Moradi, A. M. Salehoof, and S. Bibak, “ParsSimpleQA: The Persian Simple Question Answering Dataset and System over Knowledge Graph,” in Proceedings of the 2nd International Workshop on Natural Language Processing for Digital Humanities, Nov. 2022, pp. 59–68. [Online]. Available: https://aclanthology.org/2022.nlp4dh-1.9

[11] N. Jamali, Y. Yaghoobzadeh, and H. Faili, “PerCQA: Persian Community Question Answering Dataset.” 2021.

[12] Y. Boreshban, H. Yousefinasab and S A. Mirroshandel, “Providing a Religious Corpus of Question Answering System in Persian,” JSDP 2018; 15 (1) :87-102.

[13] A. Bordes, S. Chopra, and J. Weston, “Question Answering with Subgraph Embeddings,” 2014.

[14] W. T. Yih, M. Richardson, C. Meek, M. W. Chang, and J. Suh, “The value of semantic parse labeling for knowledge base question answering,” in 54th Annual Meeting of the Association for Computational Linguistics, ACL 2016 - Short Papers, 2016, pp. 201–206. doi: 10.18653/v1/p16-2033.

[15] P. Trivedi, G. Maheshwari, M. Dubey, and J. Lehmann, “LC-QuAD: A corpus for complex question answering over knowledge graphs,” in Lecture Notes in Computer Science (including subseries Lecture Notes in Artificial Intelligence and Lecture Notes in Bioinformatics), vol. 10588 LNCS, 2017, pp. 210–218. doi: 10.1007/978-3-319-68204-4_22.

[16] M. Dubey, D. Banerjee, A. Abdelkawi, and J. Lehmann, “LC-QuAD 2.0: A Large Dataset for Complex Question Answering over Wikidata and DBpedia,” in Lecture Notes in Computer Science (including subseries Lecture Notes in Artificial Intelligence and Lecture Notes in Bioinformatics), vol. 11779 LNCS, 2019, pp. 69–78. doi: 10.1007/978-3-030-30796-7_5.

[17] K. Jiang, D. Wu, and H. Jiang, “FreebaseQA: A New Factoid QA Data Set Matching Trivia-Style Question-Answer Pairs with Freebase,” in Proceedings of the 2019 Conference of the North {A}merican Chapter of the Association for Computational

Linguistics: Human Language Technologies, Volume 1 (Long and Short Papers), Jun. 2019, pp. 318–323. doi: 10.18653/v1/N19-1028.

[18] Y. Gu et al., “Beyond I.I.D.: Three levels of generalization for question answering on knowledge bases,” Web Conf. 2021 - Proc. World Wide Web Conf. WWW 2021, vol. 2021, pp. 3477–3488, 2021, doi: 10.1145/3442381.3449992.

[19] D. Biswas, M. Dubey, M. R. A. H. Rony, and J. Lehmann, “VANiLLa : Verbalized Answers in Natural Language at Large Scale.” 2021.

[20] V. Korablinov and P. Braslavski, RuBQ: A Russian Dataset for Question Answering over Wikidata, vol. 12507 LNCS. Springer International Publishing, 2020. doi: 10.1007/978-3-030-62466-8_7.

[21] Y. Han, B. Wang, R. Hong, and F. Wu, “Movie Question Answering via Textual Memory and Plot Graph,” IEEE Trans. Circuits Syst. Video Technol., vol. 30, no. 3, pp. 875–887, Mar. 2020, doi: 10.1109/TCSVT.2019.2897604.

[22] E. Kacupaj, B. Banerjee, K. Singh, and J. Lehmann, “ParaQA: A Question Answering Dataset with Paraphrase Responses for Single-Turn Conversation.” 2021.

[23] A. Abujabal, R. S. Roy, M. Yahya, and G. Weikum, “ComQA: A Community-sourced Dataset for Complex Factoid Question Answering with Paraphrase Clusters.” 2019.

[24] E. Kacupaj, H. Zafar, J. Lehmann, and M. Maleshkova, “VQuAnDa: Verbalization QUestion ANswering DAtaset,” in Lecture Notes in Computer Science (including subseries Lecture Notes in Artificial Intelligence and Lecture Notes in Bioinformatics), vol. 12123 LNCS, 2020, pp. 531–547. doi: 10.1007/978-3-030-49461-2_31.

[25] P. Christmann et al., “Look before you Hop: Conversational Question Answering over Knowledge Graphs Using Judicious Context Expansion,” in Proceedings of the 28th ACM International Conference on Information and Knowledge Management, Nov. 2019, pp. 729–738. doi: 10.1145/3357384.3358016.

[26] P. Lewis, B. Oğuz, R. Rinott, S. Riedel, and H. Schwenk, “MLQA: Evaluating Cross-lingual Extractive Question Answering.” 2020.

[27] M. Artetxe, S. Ruder, and D. Yogatama, “On the Cross-lingual Transferability of Monolingual Representations,” 2020. doi: 10.18653/v1/2020.acl-main.421.

[28] M. Joshi, E. Choi, D. S. Weld, and L. Zettlemoyer, “TriviaQA: A Large Scale Distantly Supervised Challenge Dataset for Reading Comprehension.” 2017.

[29] J. Zou et al., “A Chinese Multi-type Complex Questions Answering Dataset over Wikidata.” 2021.

[30] P. NAKOV, L. MÀRQUEZ, A. MOSCHITTI, and H. MUBARAK, “Arabic community question answering,” Nat. Lang. Eng., vol. 25, no. 1, pp. 5–41, 2019, doi: 10.1017/S1351324918000426.

[31] Z. Yang et al., “Hotpotqa: A dataset for diverse, explainable multi-hop question answering,” in Proceedings of the 2018 Conference on Empirical Methods in Natural Language Processing, EMNLP 2018, 2020, pp. 2369–2380. doi: 10.18653/v1/d18-1259.

[32] A. Conneau et al., “Unsupervised Cross-lingual Representation Learning at Scale,” 2020.

[33] L. Xue et al., “mT5: A massively multilingual pre-trained text-to-text transformer.” Conference of the North American Chapter of the Association for Computational Linguistics: Human Language Technologies ,2021.

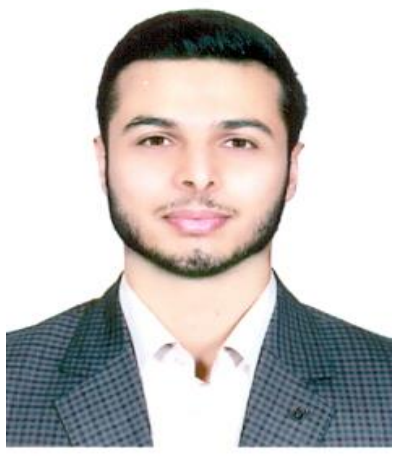

**Arash Ghafouri** Arash Ghafouri is a PhD candidate in Computer Engineering at the Iran University of Science and Technology. He is active in the field of natural language processing, information retrieval, and their applications in other areas such as search engines and chatbots. He received his Master's degree in Computer Engineering from the Iran University of Science and Technology in 2014. As an active researcher in the field of natural language processing and information retrieval, he used novel natural language processing methods and advanced machine learning techniques to solve problems related to these fields.

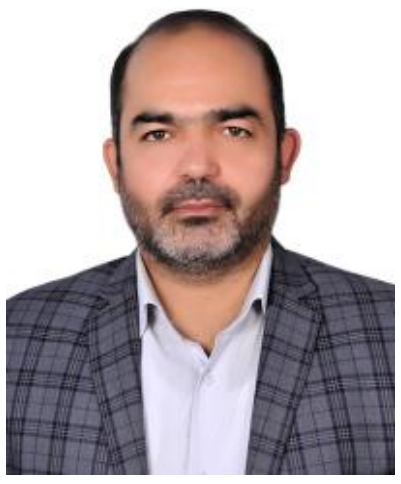

**Hassan Naderi** He is an associate professor at the Iran University of Science and Technology. He received the M.S. degree in computer engineering from the Sharif University of Technology, Tehran, in 2001, and the Ph.D. degree in information technology from the Institut National des Sciences Appliquées de Lyon, France, in 2006. He leads the Data Science and Technology Laboratory (DSTL), which researches various areas in data science and its application application in other research field.

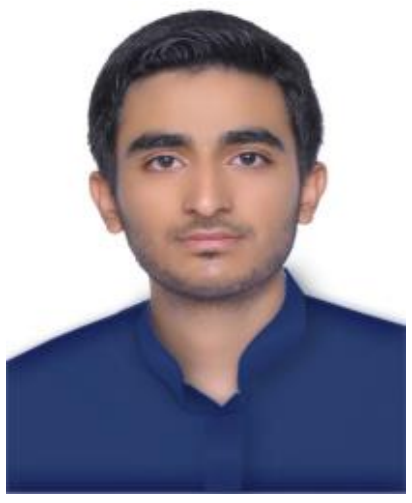

**Mohammad Aghajani Asl** received the bachelor's degree in physics from the Sharif University of Technology, Tehran, Iran, in , where he is currently pursuing the master’s degree. He is interested in research in the field of data science and text mining and is actively involved in role-playing and problem solving in this field as a researcher.

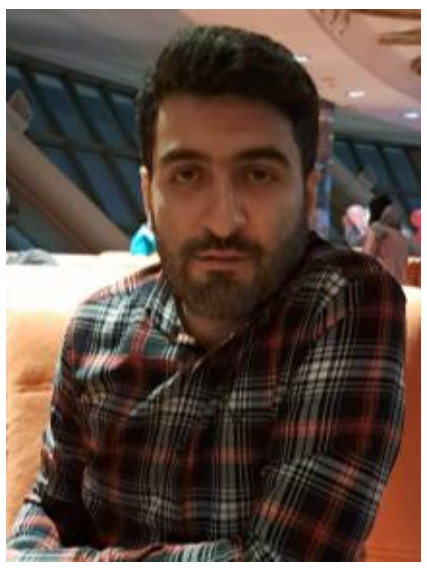

**Mahdi Firouzmandi** received the bachelor's degree in computer engineering from the Department of Computer Engineering, Iran University of Science and Technology, Tehran, Iran, in 2013, where he is currently pursuing the master’s degree. His research activity concerns artificial intelligence and machine learning, focusing on question answering system and related natural language processing (NLP) tasks. Other research interests involve deep learning and its application in other research field.